\pdfoutput=1

\documentclass[11pt]{article}
  
\usepackage[final]{acl}
\usepackage{latexsym}
\usepackage[]{natbib}

\usepackage[utf8]{inputenc}

\usepackage{microtype}

\usepackage{inconsolata}

\usepackage{graphicx}

\usepackage{caption}
\DeclareCaptionType{Algorithm}
\usepackage{amsmath}
\usepackage{algorithm}
\usepackage{algorithmic}
\usepackage{booktabs}
\usepackage{tikz}
\usepackage{float}
\usepackage{times}
\usepackage{epsfig}
\usepackage{amsmath}
\usepackage{amsfonts}
\usepackage{array}
\usepackage{multirow}
\usepackage[flushleft]{threeparttable}
\usepackage{comment}
\usepackage{algorithmic}
\usepackage{booktabs}
\usepackage{bbm, dsfont}
\usepackage[font=small,labelfont=bf]{caption}
\usepackage{subcaption}

\usepackage{makecell}

\definecolor{ForestGreen}{rgb}{0.13, 0.55, 0.13}
\definecolor{Green}{rgb}{0.0, 0.5, 0.0}
\definecolor{green(munsell)}{rgb}{0.0, 0.66, 0.47}
\definecolor{green(ryb)}{rgb}{0.4, 0.69, 0.2}
\definecolor{green(pigment)}{rgb}{0.0, 0.65, 0.31}
\definecolor{citecolor}{HTML}{0071bc}
\definecolor{GrayXMark}{gray}{0.7}

\usepackage{tabularx}
\usepackage[export]{adjustbox}
\usepackage{xcolor}

\usepackage{hyperref}
\usepackage{pifont}

\usepackage[capitalize]{cleveref}
\crefname{section}{Sec.}{Secs.}
\Crefname{section}{Section}{Sections}
\Crefname{table}{Table}{Tables}
\crefname{table}{Table}{Tabs.}

\crefname{Algorithm}{Algorithm}{Algorithms}
\Crefname{Algorithm}{Algorithm}{Algorithms}

\definecolor{ForestGreen}{rgb}{0.13, 0.55, 0.13}
\definecolor{Green}{rgb}{0.0, 0.5, 0.0}
\definecolor{green(munsell)}{rgb}{0.0, 0.66, 0.47}
\definecolor{green(ryb)}{rgb}{0.4, 0.69, 0.2}
\definecolor{green(pigment)}{rgb}{0.0, 0.65, 0.31}
\definecolor{colorlink}{RGB}{237,2,140}
\usepackage{colortbl}
\usepackage{xspace}

\usepackage{graphicx}
\usepackage{enumitem}
\usepackage{wrapfig}
\usepackage{lipsum}

\newcolumntype{H}{>{\setbox0=\hbox\bgroup}c<{\egroup}@{}}
\newcolumntype{a}{>{\columncolor{Gray}}c}

\usepackage{tabu}
\usepackage{nicematrix}

\definecolor{ForestGreen}{rgb}{0.13, 0.55, 0.13}
\definecolor{Green}{rgb}{0.0, 0.5, 0.0}
\definecolor{green(munsell)}{rgb}{0.0, 0.66, 0.47}
\definecolor{green(ryb)}{rgb}{0.4, 0.69, 0.2}
\definecolor{green(pigment)}{rgb}{0.0, 0.65, 0.31}
\definecolor{citecolor}{HTML}{0071bc}
\definecolor{GrayXMark}{gray}{0.7}

\usepackage{tabularx}
\usepackage[export]{adjustbox}

\definecolor{ForestGreen}{rgb}{0.13, 0.55, 0.13}
\definecolor{Green}{rgb}{0.0, 0.5, 0.0}
\definecolor{green(munsell)}{rgb}{0.0, 0.66, 0.47}
\definecolor{green(ryb)}{rgb}{0.4, 0.69, 0.2}
\definecolor{green(pigment)}{rgb}{0.0, 0.65, 0.31}
\definecolor{mygray}{gray}{.9}

\newcolumntype{x}[1]{>{\centering\let\newline\\\arraybackslash\hspace{0pt}}p{#1}}
\definecolor{Gray}{gray}{0.9}

\newcommand{\eg}{\textit{e.g.}\xspace}

\newcommand{\ie}{\textit{i.e.}\xspace}

\newcommand{\aka}{a.k.a.\xspace}

\definecolor{mygray}{gray}{.9}
\definecolor{colorlink}{RGB}{237,2,140}

\title{AnchorSeg: Language Grounded Query Banks for Reasoning Segmentation}

\author{
  \textbf{Rui Qian\textsuperscript{1}
  },
  \textbf{Chuanhang Deng\textsuperscript{1,2}
  },
  \textbf{Qiang Huang\textsuperscript{1,2}
  },
  \textbf{Jian Xiong\textsuperscript{1}},\\
  \textbf{Mingxuan Li\textsuperscript{1}}, 
  \textbf{Yingbo Zhou\textsuperscript{1}}, 
  \textbf{Wei Zhai\textsuperscript{1}},
  \textbf{Jintao Chen\textsuperscript{1}},
  \textbf{Dejing Dou\textsuperscript{1,2}\thanks{Corresponding author}} 
\\
\textsuperscript{1}College of Computer Science and Artificial Intelligence, Fudan University \quad \textsuperscript{2}BEDI Cloud \\
\texttt{\{qiianruii,dengch2000,dejingdou\}@gmail.com} \\
}
   
\begin{document}
\maketitle
\begin{abstract}
    Reasoning segmentation requires models to ground complex, implicit textual queries 
    into precise pixel-level masks. Existing approaches 
    rely on a single segmentation token \texttt{<SEG>}, whose hidden state implicitly 
    encodes both semantic reasoning and spatial localization, limiting the model's ability to 
    explicitly disentangle \emph{what to segment} from \emph{where to segment}. We 
    introduce AnchorSeg, which reformulates reasoning 
    segmentation as a structured conditional generation process over image tokens, 
    conditioned on language grounded query banks. Instead of compressing both semantic 
    reasoning and spatial localization into a single embedding, AnchorSeg constructs an ordered sequence of query 
    banks: latent reasoning tokens that capture intermediate semantic states, 
    and a segmentation anchor token that provides explicit spatial grounding. We model 
    spatial conditioning as a factorized distribution over image tokens, where the 
    anchor query determines localization signals while contextual queries provide 
    semantic modulation. To bridge token-level predictions and pixel-level supervision, 
    we propose Token--Mask Cycle Consistency (TMCC), a bidirectional training objective 
    that enforces alignment across resolutions. By explicitly decoupling spatial 
    grounding from semantic reasoning through structured language grounded 
    query banks, AnchorSeg achieves state-of-the-art results on ReasonSeg test set 
    (67.7\% gIoU and 68.1\% cIoU). 
    All code and models are publicly available at {\hypersetup{urlcolor=colorlink}\href{https://github.com/rui-qian/AnchorSeg}{https://github.com/rui-qian/AnchorSeg}}.
\end{abstract}

\section{Introduction}
\label{sec: intro}

Reasoning segmentation aims to predict pixel-level segmentation masks by grounding 
complex language reasoning into visual scenes~\citep{lai2024lisa}.
Unlike traditional referring segmentation, which operates on explicit object 
descriptions (\eg, ``\emph{the red car on the left}''), 
reasoning segmentation involves reasoning over abstract concepts, spatial relationships, 
and implicit references (\eg, ``\emph{the object that provides shade in this scene}'') 
to identify and segment target regions. Such capability is essential for intelligent 
vision–language systems that can accurately interpret and respond to human intent 
in complex visual environments.

Recent advances in Large Multimodal Models (LMMs) have demonstrated 
remarkable progress in combining language understanding with visual 
perception~\citep{liu2023llava,li2024llavanext-ablations,alayrac2022flamingo,wang2023visionllm,
zhu2023minigpt,zhang2023gpt4roi}. To endow LMMs with segmentation capability, 
LISA~\citep{lai2024lisa} introduces a special segmentation token \texttt{<SEG>} into 
the language model vocabulary. During autoregressive generation, when the model produces 
this token, its hidden representation is extracted and used as a single, unified query 
to condition a visual decoder (\eg, SAM~\citep{kirillov2023segment}) for mask prediction. 
While effective, this paradigm~\citep{wu2023see,xia2024gsva,qian2024reasoning} compresses both semantic reasoning and spatial localization 
cues into a single \texttt{<SEG>} embedding vector. Such implicit compression limits 
the model's ability to explicitly disentangle \emph{what to segment} (semantic reasoning) from 
\emph{where to segment} (spatial grounding), potentially hindering the model's performance 
in complex reasoning scenarios.

We argue that reasoning segmentation can be formulated as a structured conditional 
generation process, where spatial grounding is performed at the image-token level and 
explicitly conditioned on language-derived queries that encode intermediate reasoning 
states. To this end, we propose AnchorSeg, a framework built upon language 
grounded query banks, where an explicit \textbf{Anchor} query grounds language reasoning 
to visual tokens and guides \textbf{Seg}mentation. Instead of relying on a single 
segmentation token, AnchorSeg introduces a set of learnable latent reasoning tokens 
alongside a segmentation anchor token, which are autoregressively generated by LMMs. This 
design yields an ordered sequence of query embeddings: contextual queries that capture 
intermediate reasoning states, and an anchor query that serves as an explicit spatial grounding signal.

Our key innovation lies in how these queries interact with visual representations. 
We model spatial grounding as a factorized conditional distribution over image tokens, 
where each token's relevance to the target object is  conditioned on both the 
contextual queries (providing semantic modulation) and the anchor query (determining 
spatial grounding). This formulation enables explicit token-level language grounding, 
producing a spatial prior that is injected into the visual features before mask decoding. 
The entire query banks, ordered by the autoregressive generation process, are then fed into 
the SAM decoder, providing structured, language-conditioned supervision for 
final mask prediction.

To bridge the gap between token-level spatial responses and pixel-level mask supervision, 
we introduce a Token--Mask Cycle Consistency (TMCC) training objective. This 
bidirectional constraint enforces alignment between the language grounded token-level 
predictions and the ground-truth masks at both resolutions, ensuring that spatial 
reasoning is consistent across the language-vision hierarchy. By explicitly modeling 
token-level spatial response factorization through structured language grounded 
query banks, AnchorSeg establishes a more effective coupling between language reasoning 
and visual segmentation.

Our contributions can be summarized as follows:
\begin{itemize}[leftmargin=*, itemsep=2pt, parsep=0pt, topsep=4pt]
    \item We reformulate reasoning segmentation as a structured conditional generation 
    problem, introducing language grounded query banks that explicitly disentangle semantic reasoning from spatial grounding at the image-token level.
    \item We propose a factorized formulation for language grounded  conditioning, 
    where anchor queries produce explicit localization signals and contextual queries 
    provide semantic modulation, enabling enhanced language-to-vision alignment.
    \item We introduce Token--Mask Cycle Consistency (TMCC), a bidirectional training objective that enforces alignment between token-level spatial responses and pixel-level mask supervision across resolutions.
    \item We conduct extensive experiments on ReasonSeg and RefCOCO(+/g) datasets. The 
    proposed AnchorSeg achieves state-of-the-art results on 
    ReasonSeg test set (67.7\% gIoU and 68.1\% cIoU).
\end{itemize}

\section{Related Work}
\label{sec: related_work}

\subsection{Large Multimodal Models}
Large Multimodal Models (LMMs) underpin reasoning segmentation by enabling joint perception and language 
understanding. Representative models include VisionLLM~\citep{wang2023visionllm}, MiniGPT-4~\citep{zhu2023minigpt}, and GPT4RoI~\citep{zhang2023gpt4roi}, which show that integrating LLMs with visual features enables open-ended multimodal reasoning, particularly in region-level grounding and object-level understanding. BLIP~\citep{li2023blip2} focuses on vision–language alignment by leveraging frozen image encoders while supporting generative multimodal modeling. Flamingo~\citep{alayrac2022flamingo} introduces a flexible interleaved vision–text interface that enables few-shot multimodal reasoning across diverse vision–language tasks. LLaVA~\citep{liu2023llava} and LLaVA-NeXT~\citep{li2024llavanext-ablations} further enhance visual reasoning and document understanding, while more recent works~\citep{hurst2024gpt, wu2024next, gpt5} integrate text, audio, and video into unified multimodal models.

Despite existing efforts, current LMMs still primarily generate textual responses and struggle to deliver structured dense predictions, often relying on weak or implicit grounding signals rather than precise pixel-level masks~\citep{radford2021learningclip, li2023blip2, alayrac2022flamingo}. Recent works extend LMMs toward pixel-level visual grounding by introducing segmentation tokens, pixel decoders, visual prompting modules and chain-of-thought reasoning for grounding~\citep{lai2024lisa, ren2024pixellm, qian2024reasoning}. In contrast, our approach introduces a structured formulation that explicitly bridges language reasoning and pixel-level segmentation, enabling more interpretable grounding.

\subsection{Reasoning Segmentation}
Reasoning segmentation extends referring expression segmentation to settings where models are 
required to infer implicit visual concepts and output pixel-level masks under complex scenarios.
LISA~\citep{lai2024lisa} pioneers this task, injecting a \texttt{<SEG>} 
token into LMMs and decoding its embedding into masks under an ``embedding-as-mask'' paradigm, 
accompanied by the ReasonSeg benchmark for implicit reasoning queries.
Follow-up works expand this formulation: GSVA~\citep{xia2024gsva} generalizes the 
segmentation token to multi-target reasoning and introduces \texttt{<REJ>} for non-existent-object 
rejection. SESAME~\citep{wu2023see} mitigates hallucination and false-premise failures 
via a pipeline that verifies existence, overcomes the premise, and then segments. 
READ~\citep{qian2024reasoning} looks into how \texttt{<SEG>} contributes 
to grounding and proposes a ``similarity as points'' module for enhancement.

Beyond token design, recent research explores pixel-grounded LMMs with richer 
interactivity. GLaMM~\citep{rasheed2024glamm} enables joint language generation and 
segmentation in grounded conversations, while PixelLM~\citep{ren2024pixellm} integrates 
a lightweight pixel decoder and segmentation codebook directly into LMMs, 
removing reliance on the external SAM~\citep{ravi2024sam}.  RSVP~\citep{lu2025rsvp} further couples multimodal chain-of-thought reasoning with segmentation, generating interpretable 
region proposals before refinement. In this work, we model reasoning segmentation as 
token-level conditional generation of spatial responses, factorized over image tokens 
and conditioned on a structured language query bank.

\begin{figure*}[h]
\begin{center}
\includegraphics[width=1\linewidth]{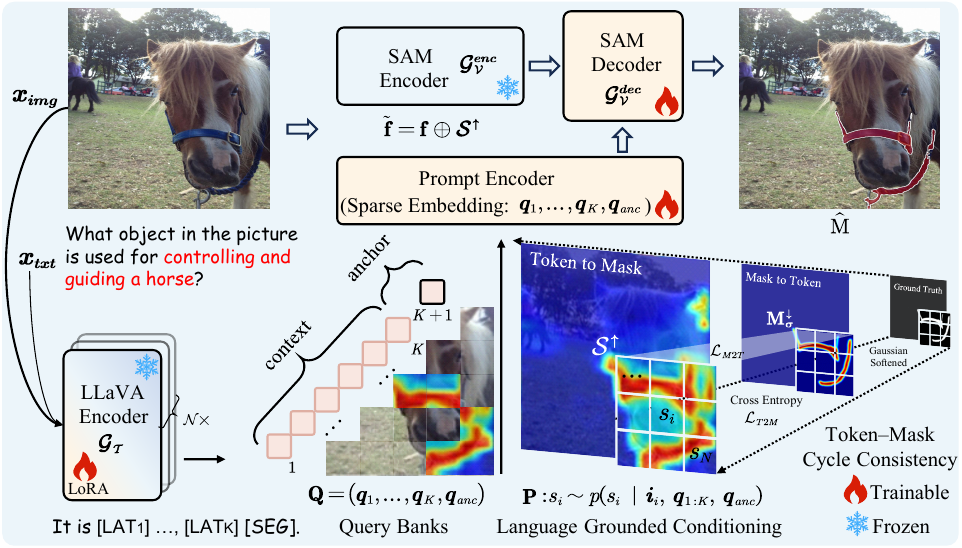}
\end{center}
\vspace{-0.4cm}
\caption{\textbf{Overview of AnchorSeg.} Given an input image and textual query, the LMM $\mathcal{G}_{\mathcal{T}}$  autoregressively generates latent reasoning tokens and a segmentation anchor token \texttt{<SEG>}, forming a language grounded query bank $\mathbf{Q} = \{\boldsymbol{q}_1, \dots, \boldsymbol{q}_K, \boldsymbol{q}_{anc}\}$. The anchor query $\boldsymbol{q}_{anc}$ computes similarity with image tokens to produce a spatial prior $\mathbf{P}$, which is injected into visual features $\mathbf{f}$. The query bank then conditions the SAM decoder to predict the final mask $\hat{\mathbf{M}}$. This design explicitly disentangles spatial grounding (anchor query) from semantic reasoning (contextual queries).}
\label{fig:method}
\vspace{-0.4cm}
\end{figure*}

\section{Proposed AnchorSeg}
\label{sec:reasoning_seg}

In this section, we present AnchorSeg, which formulates reasoning segmentation as 
conditional decoding by jointly leveraging language grounded spatial priors and 
structured query conditioning. In Fig.~\ref{fig:method}, AnchorSeg comprises three key components: 
1) an LMM $\mathcal{G}_{\mathcal{T}}$ that autoregressively generates latent reasoning tokens and a segmentation anchor token \texttt{<SEG>}, forming a language grounded query bank $\mathbf{Q}$; 
2) a language grounded spatial conditioning module that computes similarity between the anchor query and image tokens to produce a spatial prior $\mathbf{P}$; 3) a SAM mask decoder conditioned on the query bank to predict the final segmentation mask. Given an input image and a textual prompt, AnchorSeg 
generates a \texttt{<SEG>} token to indicate the target object, along with a set 
of latent reasoning tokens that encode intermediate semantic cues. 
Unlike prior works, AnchorSeg explicitly grounds the \texttt{<SEG>} token to 
image tokens via a similarity map, producing a coarse spatial localization signal. 
Instead of using a single \texttt{<SEG>} token, we elevate this signal into a 
language grounded spatial prior and inject it into the visual feature space, 
enabling segmentation to be guided by both explicit visual priors and structured 
language queries.

\subsection{Vanilla Reasoning Segmentation with Language-Conditioned \texttt{<SEG>} Query}
\label{sec:vanilla_reason_seg}

\paragraph{Problem Definition:} We first revisit the vanilla formulation of reasoning segmentation 
proposed in LISA~\citep{lai2024lisa}. Given an input image $\mathbf{x}_{img} \in \mathbb{R}^{h \times w \times c}$ and a textual prompt $\mathbf{x}_{txt}$,
the goal of reasoning segmentation is to predict a binary mask
$\hat{\mathbf{M}} \in \{0,1\}^{h \times w}$ corresponding to the visual concept described 
in the text as
\begin{equation}
\hat{\mathbf{M}}
=
\arg\max_{\hat{\mathbf{M}}}
\mathcal{G}_{\theta}
\big(
\hat{\mathbf{M}}
\mid
\mathbf{x}_{img}, \mathbf{x}_{txt}
\big),
\end{equation}
where $\mathcal{G}_{\theta} = \mathcal{G}_{\mathcal{T}} \oplus \mathcal{G}_{\mathcal{V}}$
denotes a cascaded architecture composed of an LMM $\mathcal{G}_{\mathcal{T}}$ (\eg,
LLaVA~\citep{liu2023llava}), and a visual backbone model $\mathcal{G}_{\mathcal{V}}$ 
(\eg, SAM~\citep{kirillov2023segment}).

To enable segmentation within a language modeling framework,
the vanilla approach extends the vocabulary of LMM $\mathcal{G}_{\mathcal{T}}$
with a special placeholder, denoted as \texttt{<SEG>}.
During autoregressive generation,
the language model produces an output token sequence as
\begin{equation}
\hat{\mathbf{y}}_{txt}
=
\mathcal{G}_{\mathcal{T}}(\mathbf{x}_{img}, \mathbf{x}_{txt}),
\end{equation}
in which the \texttt{<SEG>} token is generated when a segmentation output is required. 
Let $\boldsymbol{q}_{seg}$ denote the hidden representation
corresponding to the predicted \texttt{<SEG>} token
extracted from the final transformer layer
and projected via a lightweight MLP $\varphi(\cdot)$, which serves as a single,
language-conditioned segmentation query that implicitly encodes both
semantics and spatial guidance. On the visual side, the segmentation model
$\mathcal{G}_{\mathcal{V}}^{enc}$ (instantiated by SAM~\citep{kirillov2023segment})
first encodes image $\mathbf{x}_{img}$ into SAM's dense visual features as
\begin{equation}
\boldsymbol{q}_{seg}
=
\varphi(\tilde{\boldsymbol{h}}_{seg})
\in
\mathbb{R}^d, \quad
\mathbf{f}
=
\mathcal{G}_{\mathcal{V}}^{enc}(\mathbf{x}_{img}).
\end{equation}
The final segmentation mask is then predicted by conditioning
the SAM mask decoder on the single segmentation query:
\begin{equation}
\hat{\mathbf{M}}
=
\mathcal{G}_{\mathcal{V}}^{dec}(\mathbf{f}, \boldsymbol{q}_{seg}).
\end{equation}
In this vanilla paradigm,
the \texttt{<SEG>} token serves as a single, unified interface
between language reasoning and visual segmentation. Both semantic reasoning and spatial localization cues
are implicitly compressed into a single segmentation query, 
without preserving the internal structure of the language reasoning process.

\subsection{Extended Reasoning Segmentation with Language Grounded Query Banks}
\label{sec:extended_reason_seg}

While effective, the vanilla formulation relies on a single segmentation token
to implicitly bridge language reasoning and mask decoding.
The alignment between linguistic semantics and visual representations
remains largely unexplored. We therefore reformulate reasoning segmentation
as a conditional decoding problem, where spatial grounding is explicitly modeled 
at the image-token level and jointly conditioned on structured language queries. 

\paragraph{Query Bank Construction.}
We construct a conditional query sequence by extending the vocabulary of LMM $\mathcal{G}_{\mathcal{T}}$ with a set of learnable placeholders.
Specifically, we introduce $K$ latent reasoning tokens together with a segmentation 
token \texttt{<SEG>}, which are inserted into the textual response and produced 
autoregressively as part of the language modeling process, such that the segmentation 
token \texttt{<SEG>} is explicitly conditioned on the preceding latent reasoning 
tokens, yielding the ordered query sequence as
${\texttt{<LAT}_1\texttt{>}, \dots, \texttt{<LAT}_K\texttt{>}, \texttt{<SEG>}}$.
Rather than operating on discrete symbols, we formulate the conditional query sequence
in the continuous embedding space.
Let the final-layer hidden states corresponding to the latent reasoning tokens
and the segmentation token be denoted as
$\boldsymbol{q}_1, \dots, \boldsymbol{q}_K \in \mathbb{R}^d$
and
$\boldsymbol{q}_{anc} \in \mathbb{R}^d$, respectively.
We define the conditional query sequence as
\begin{equation}
    \mathbf{Q}
    =
    \big(
    \boldsymbol{q}_1, \dots, \boldsymbol{q}_K, \boldsymbol{q}_{anc}
    \big),
\end{equation}
where the contextual queries $\boldsymbol{q}_{1:K}$ encode intermediate reasoning states
that condition the segmentation process at the semantic level,
while the anchor query $\boldsymbol{q}_{anc}$ serves as a spatial grounding token.
This ordered query sequence provides structured, language-conditioned supervision
for subsequent spatial conditioning and mask decoding.

\paragraph{Language Grounded Conditioning.}
Inspired by token-level factorization in autoregressive language modeling,
we formulate language grounded spatial conditioning
as a conditional generation process over image tokens.
Let $\boldsymbol{I} = \{\boldsymbol{i}_1, \dots, \boldsymbol{i}_N\}$,
$\boldsymbol{i}_i \in \mathbb{R}^d$,
denote the image token representations produced by LMM $\mathcal{G}_{\mathcal{T}}$.
We define a set of spatial responses
$\boldsymbol{S} = \{s_1, \dots, s_N\}$,
where each $s_i$ corresponds to the relevance of image token $\boldsymbol{i}_i$
to the target object.
The conditional distribution of $\boldsymbol{S}$ given the query bank
is factorized as
\begin{equation}
p(\boldsymbol{S} \mid \mathbf{Q})
=
\prod_{i=1}^{N}
p\!\left(
s_i
\;\middle|\;
\boldsymbol{i}_i,\;
\underbrace{\boldsymbol{q}_1,\dots,\boldsymbol{q}_K}_{\text{context}},\;
\underbrace{\boldsymbol{q}_{anc}}_{\text{anchor}}
\right).
\label{eq:token_factorization}
\end{equation}
This formulation explicitly disentangles the roles of the query tokens:
the anchor query $\boldsymbol{q}_{anc}$ determines where to attend in the image,
while the contextual queries $\boldsymbol{q}_{1:K}$ provide semantic modulation
under which spatial grounding is performed.

Under the formulation in Eq.~\eqref{eq:token_factorization},
each spatial response $s_i$ is generated conditionally
from the corresponding image token $\boldsymbol{i}_i$
and the language grounded query bank,
$s_i \sim p\!\left(s_i \;\middle|\;\boldsymbol{i}_i,\;\boldsymbol{q}_{1:K},\;\boldsymbol{q}_{anc}\right)$.
Practically, we instantiate the conditional distribution
using an anchor-based similarity function as
\begin{equation}
s_i = \boldsymbol{i}_i^{\top} \boldsymbol{q}_{anc},
\end{equation}
where the anchor query explicitly produces a localization signal over image tokens.
Although the similarity computation depends only on $\boldsymbol{q}_{anc}$,
the contextual queries $\boldsymbol{q}_{1:K}$ influence each spatial response $s_i$
implicitly by shaping the generation of the anchor query through autoregressive reasoning. 
The token-level responses $\boldsymbol{S}$ are then reshaped and projected
to the spatial domain to obtain a language grounded spatial prior
$\mathbf{P} \in \mathbb{R}^{C \times H \times W}$. Next, this spatial prior is injected into the visual feature map
$\mathbf{f} \in \mathbb{R}^{C \times H \times W}$
via element-wise addition, yielding conditioned visual features
$\tilde{\mathbf{f}} = \mathbf{f} \oplus \mathbf{P}$.

\paragraph{Conditional Mask Decoding.}
When interacting with the SAM~\citep{kirillov2023segment} mask decoder,
the language grounded query bank is treated as an ordered query sequence.
Specifically, the query sequence
$\{\boldsymbol{q}_1, \dots, \boldsymbol{q}_K, \boldsymbol{q}_{anc}\}$
is augmented with learnable positional embeddings
$\{\boldsymbol{p}_1, \dots, \boldsymbol{p}_{K+1}\}$
prior to cross-attention, ensuring both order-awareness
and explicit role distinction between contextual and anchor queries.
The position-augmented query bank is then used to condition the SAM mask decoder,
which predicts the final segmentation mask as
\begin{equation}
\hat{\mathbf{M}}
=
\mathcal{G}_{\mathcal{V}}^{dec}
\big(
\tilde{\mathbf{f}},\;
\{\boldsymbol{q}_1, \dots, \boldsymbol{q}_K, \boldsymbol{q}_{anc}\}
\big),
\end{equation}
where mask prediction is explicitly conditioned
on both language grounded spatial prior $\mathbf{P}$ and structured 
reasoning queries $(\boldsymbol{q}_{1:K}, \boldsymbol{q}_{anc})$. This formulation explicitly decouples 
spatial grounding and semantic reasoning. The segmentation anchor query 
determines spatial correspondence, while the latent reasoning queries provide 
contextual modulation. By unifying both components under a conditional decoding framework,
AnchorSeg enables robust reasoning-based segmentation guided by
structured language representations and explicit visual priors.

\subsection{Training Objectives}
\label{TrainingObjectives}

\noindent\textbf{Token--Mask Cycle Consistency (TMCC).}
Under the unified formulation in Eq.~\eqref{eq:token_factorization},
the spatial responses $\boldsymbol{S}$ act as a latent representation
bridging token-level language grounding and pixel-level mask supervision.
To regularize this representation across resolutions,
we introduce a token--mask cycle consistency objective.
Given a ground-truth binary mask $\mathbf{M} \in \{0,1\}^{H \times W}$,
we apply Gaussian smoothing to obtain a softened target mask
$\mathbf{M}_{\sigma} \in [0,1]^{H \times W}$. 1) Token-to-Mask Consistency.
We reshape and upsample the token-level responses $\boldsymbol{S}$
to the image resolution, yielding a spatial map
$\mathcal{S}^{\uparrow} \in [0,1]^{H \times W}$, which is supervised by 
$\mathbf{M}_{\sigma}$ using binary cross-entropy and Dice losses as
\begin{equation}
\mathcal{L}_{T2M}
=
\lambda_{bce}\mathcal{L}_{bce}(\mathcal{S}^{\uparrow}, \mathbf{M}_{\sigma})
+
\lambda_{dice}\mathcal{L}_{dice}(\mathcal{S}^{\uparrow}, \mathbf{M}_{\sigma}).
\end{equation} 2) Mask-to-Token Consistency.
Conversely, we downsample the soft target mask $\mathbf{M}_{\sigma}$
to the image-token resolution,
obtaining $\mathbf{M}_{\sigma}^{\downarrow} \in [0,1]^N$,
and enforce token-level alignment with $\boldsymbol{S}$ as
\begin{equation}
\mathcal{L}_{M2T}
=
\lambda_{bce}\mathcal{L}_{bce}(\boldsymbol{S}, \mathbf{M}_{\sigma}^{\downarrow})
+
\lambda_{dice}\mathcal{L}_{dice}(\boldsymbol{S}, \mathbf{M}_{\sigma}^{\downarrow}).
\end{equation}

\noindent\textbf{Overall Objectives.}
The two terms form a bidirectional cycle consistency constraint
over the latent spatial responses $\boldsymbol{S}$ as
\begin{equation}
\mathcal{L}_{TMCC} = \mathcal{L}_{T2M} + \mathcal{L}_{M2T}.
\end{equation}
The final training objective combines TMCC
with the language modeling loss $\mathcal{L}_{txt}$
and the segmentation loss $\mathcal{L}_{mask}$ as
\begin{equation}
\mathcal{L}
=
\lambda_{txt}\mathcal{L}_{txt}
+
\lambda_{mask}\mathcal{L}_{mask}
+
\lambda_{TMCC}\mathcal{L}_{TMCC}.
\end{equation}

\begin{table*}[htbp]
    \centering
    \caption{
    Performance comparison of reasoning segmentation models on the ReasonSeg dataset. Models are sorted by cIoU scores. * denotes results reproduced from official implementations, and ft indicates models fine-tuned on 239 samples.
}
    \vspace{-0.1cm}\resizebox{\linewidth}{!}
    {
    \label{table:reason_seg}   
    \tabcolsep=0.35cm
    {
        \begin{tabular}{l|cc|cc|cc|cc}
            \toprule
            
            \multirow{3}*{Method} & \multicolumn{2}{c|}{val} & \multicolumn{6}{c}{test} \\ 
            
            \specialrule{0em}{0pt}{1pt}
            \cline{2-9}
            \specialrule{0em}{0pt}{1pt}
                        
            ~ & \multicolumn{2}{c|}{overall} & \multicolumn{2}{c|}{short query} & \multicolumn{2}{c|}{long query} & \multicolumn{2}{c}{overall} \\

            \specialrule{0em}{0pt}{1pt}
            \cline{2-9}
            \specialrule{0em}{0pt}{1pt}
            
            ~ & gIoU & cIoU & gIoU & cIoU & gIoU & cIoU & gIoU & cIoU \\ 
            
            \specialrule{0em}{0pt}{1pt}
            \hline
            \specialrule{0em}{0pt}{1pt}
            X-Decoder~\cite{zou2023generalized} & 22.6 & 17.9 & 20.4 & 11.6 & 22.2 & 17.5 & 21.7 & 16.3 \\

            Grounded-SAM~\cite{liu2023grounding} & 26.0 & 14.5 & 17.8 & 10.8 & 22.4 & 18.6 & 21.3 & 16.4 \\

            SEEM~\cite{zou2024seem} & 25.5 & 21.2 & 20.1 & 11.5 & 25.6 & 20.8 & 24.3 & 18.7 \\
            
            OVSeg~\cite{liang2023open} & 28.5 & 18.6 & 18.0 & 15.5 & 28.7 & 22.5 & 26.1 & 20.8  \\

            GRES~\cite{liu2023gres} & 22.4 & 19.9 & 17.6 & 15.0 & 22.6 & 23.8 & 21.3 & 22.0 \\    %
    
            \specialrule{0em}{0pt}{1pt}
            \hline
            \specialrule{0em}{0pt}{1pt}
            
            *SESAME~\cite{wu2023see} & {40.3} & {41.6} & {28.9} & {26.3} & {37.3} & {31.9}& {34.9} & {30.7} \\

            LLaVA1.5-7B + OVSeg~\cite{lai2024lisa} & 38.2 & 23.5 & 24.2 & 18.7 & 44.6 & 37.1 & 39.7 & 31.8 \\

            *GSVA~\cite{xia2024gsva} & {45.6} & {41.5} & {37.9} & {36.5} & {44.3} & {46.0}& {42.8} & {43.8} \\

            *PixelLM~\cite{ren2024pixellm} & {49.7} & {49.6} & {39.5} & {38.8} & {49.5} & {45.6}& {47.1} & {44.3} \\

            LISA-7B~\cite{lai2024lisa} & 52.9 & 54.0 & 40.6 & 40.6 & 49.4 & 51.0 & 47.3 & 48.4 \\

            HyperSeg-3B~\cite{wei2025hyperseg} & 59.2 & 56.7 & - & - & - & - & - & - \\

            VISA-7B~\cite{yan2024visa} & 52.7 & 57.8 & - & - & - & - & - & - \\

            VideoLISA-3.8B~\cite{bai2024one} & 61.4 & 67.1 & 43.8 & 42.7 & 56.9 & 57.7 & 53.8 & 54.4 \\

            LISA-7B-LLaVA1.5 (ft)~\cite{lai2024lisa} & 61.3 & 62.9 & {48.3} & {46.3} & 57.9 & 59.7 & 55.6 & 56.9 \\
            
            READ-7B-LLaVA1.5 (ft)~\cite{qian2024reasoning} & 59.8 & 67.6 & 52.6 & 49.5  & 60.4 & 61.0 & 58.5 & 58.6 \\
            
            LISA++-7B-LLaVA1.5 (ft)~\cite{yang2023lisa++} & 64.2 & 68.1 & {49.6} & \textbf{51.1} & 59.3 & 61.7 & 57.0 & 59.5 \\

            RSVP-GPT~\cite{lu2025rsvp} & 64.7 & 63.1 & {55.4} & 50.4 & 61.9 & 62.5 & 60.3 & 60.0 \\
            \rowcolor{mygray}
            AnchorSeg-7B-LLaVA1.5 (ft) & \textbf{68.3} & \textbf{75.9} & \textbf{57.3} & 48.2  & \textbf{67.0} & \textbf{71.3} & \textbf{64.6} & \textbf{65.9} \\
           \specialrule{0em}{0pt}{1pt}
            \hline 
            Qwen3-VL-8B~\cite{liang2026seg} & {70.3} & {70.0} & - & -  & - & - & {66.0} & {53.7} \\
            Seg-ReSearch-8B~\cite{liang2026seg} & \textbf{73.3} & {72.2} & - & -  & - & - & {67.4} & {59.0} \\

            LLaVA1.5-13B + OVSeg~\cite{lai2024lisa} & 37.9 & 26.4 & 27.1 & 19.4 & 46.1 & 40.6 & 41.5 & 34.1 \\
            
            LISA-13B-LLaVA1.5~\cite{lai2024lisa} & 57.7 & 60.3 & 50.8 & 50.0 & 54.7 & 50.9 & 53.8 & 50.8 \\
               
            LISA-13B-LLaVA1.5(ft)~\cite{lai2024lisa} & 65.0 & 72.9 & 55.4 & 50.6 & 63.2 & 65.3 & 61.3 & 62.2 \\

            READ-13B-LLaVA1.5 (ft)~\cite{qian2024reasoning} & - & - & 55.4 & \textbf{53.7}  & 64.4 & 65.1 & 62.2 & 62.8 \\
            \rowcolor{mygray}
            AnchorSeg-13B-LLaVA1.5 (ft) & {67.9} & \textbf{73.0} & \textbf{59.6} & 51.8 & \textbf{70.2} & \textbf{73.3} & \textbf{67.7} & \textbf{68.1} \\
        
            \bottomrule            
        \end{tabular}
    }  
    }  
\end{table*}

\begin{table*}[htbp]
    \centering 
    \caption{
    Results for referring expression comprehension on RefCOCO, RefCOCO+, and RefCOCOg (Precision@0.5). (full-ft) denotes full training of the u-LLaVA LLM. For u-LLaVA-7B, we report results using the “mask2bbox” strategy to ensure fair comparison.
    } \vspace{-0.2cm}\resizebox{\linewidth}{!}
    {
    \label{table:refer_seg_c}   
    \tabcolsep=0.4cm
    {
        \begin{tabular}{l|ccc|ccc|cc}
            \toprule
            
            \multirow{2}*{Method} & \multicolumn{3}{c|}{RefCOCO} & \multicolumn{3}{c|}{RefCOCO+}  & \multicolumn{2}{c}{RefCOCOg} \\ 
            
            \specialrule{0em}{0pt}{1pt}
            \cline{2-9}
            \specialrule{0em}{0pt}{1pt}
            
            ~ & val & testA & testB & val & testA & testB & val & test \\ 
                 
            \specialrule{0em}{0pt}{1pt}
            \hline
            \specialrule{0em}{0pt}{1pt}

            u-LLaVA-7B~\cite{xu2023u} (full-ft) & {86.04} & {89.47} & {82.26} & 74.09 & 81.16 & {66.61} & {79.87} & {81.68} \\

            LISA-Vicuna-7B~\cite{lai2024lisa} (ft) & {85.39} & {88.84} & {82.59} & 74.23 & 79.46 & {68.40} & {79.34} & {80.42} \\

            GSVA-Vicuna-7B~\cite{xia2024gsva} (ft) & {86.27} & {89.92} & {83.77} & 72.81 & 78.78 & {68.01} & {81.58} & {81.83} \\
            
            
            \rowcolor{mygray}
            AnchorSeg-LLaVA1.5-7B & \textbf{89.10} & \textbf{92.10} & \textbf{84.49} & \textbf{80.89} & \textbf{85.87} & \textbf{73.29} & \textbf{84.09} & \textbf{84.45} \\
                \specialrule{0em}{0pt}{1pt} \hline \specialrule{0em}{0pt}{1pt}

    LISA-Vicuna-13B~\cite{lai2024lisa} (ft) & {85.92} & {89.05} & {83.16} & 74.86 & 81.08 & {68.87} & {80.09} & {81.48} \\

    GSVA-Vicuna-13B~\cite{xia2024gsva} (ft) & {87.71} & {90.49} & {84.57} & 76.52 & 81.69 & {70.35} & {83.90} & {84.85} \\


    \rowcolor{mygray} AnchorSeg-LLaVA1.5-13B & \textbf{91.34} & \textbf{94.56} & \textbf{87.54} & \textbf{84.25} & \textbf{89.75} & \textbf{77.71} & \textbf{87.79} & \textbf{87.09} \\
            \bottomrule            
        \end{tabular}
    }
    }
\end{table*}
\begin{table*}[htbp]
    \centering
    \caption{
    Benchmarking Generalized Referring Expression Segmentation (GRES) on the gRefCOCO dataset~\cite{liu2023gres}. Models are listed in ascending order based on the cIoU of the \textit{val} set. Values are derived from~\cite{liu2023gres}. N-acc. denotes the accuracy of identifying null targets, and ft indicates models fine-tuned on the gRefCOCO training split.
    }
    \vspace{-0.2cm}
    \tabcolsep=0.3cm
    \resizebox{\linewidth}{!}{
    \begin{tabular}{l|ccc|ccc|ccc}
    \toprule
    \multirow{2}{*}{Method} & \multicolumn{3}{c|}{Validation Set} & \multicolumn{3}{c|}{Test Set A}  & \multicolumn{3}{c}{Test Set B} \\
    \specialrule{0em}{0pt}{1pt}
    \cline{2-10}
    \specialrule{0em}{0pt}{1pt}
    & gIoU & cIoU & N-acc. & gIoU & cIoU & N-acc. & gIoU & cIoU & N-acc. \\
    \midrule
    MattNet~\cite{yu2018mattnet} & 48.24 & 47.51 & 41.15 & 59.30 & 58.66 & 44.04 & 46.14 & 45.33 & 41.32 \\
    LTS~\cite{jing2021locate} & 52.70 & 52.30 & - & 62.64 & 61.87 & - & 50.42 & 49.96 & - \\
    VLT~\cite{ding2021vision} & 52.00 & 52.51 & 47.17 & 63.20 & 62.19 & 48.74 & 50.88 & 50.52 & 47.82 \\
    CRIS~\cite{wang2022cris} & 56.27 & 55.34 & - & 63.42 & 63.82 & - & 51.79 & 51.04  & - \\
    LAVT~\cite{yang2022lavt} & 58.40 & 57.64 & 49.32 & 65.90 & 65.32 & 49.25 & 55.83 & 55.04 & 48.46 \\
    ReLA~\cite{liu2023gres} & 63.60 & 62.42 & 56.37 & 70.03 & 69.26 & 59.02 & 61.02 & 59.88 & 58.40 \\
    \midrule
    LISA-Vicuna-7B~\cite{lai2024lisa} & 32.21 & 38.72 & 2.71 & 48.54 & 52.55 & 6.37 & 39.65 & 44.79 & 5.00 \\
    GSVA-Vicuna-7B~\cite{xia2024gsva} & 63.32 & 61.70 & 56.45 & 70.11 & 69.23 & 63.50 & 61.34 & 60.26 & 58.42  \\
    LISA-Vicuna-7B (ft)~\cite{lai2024lisa} & 61.63 & 61.76 & 54.67 & 66.27 & 68.50 & 50.01 & 58.84 & 60.63 & 51.91 \\
    GSVA-Vicuna-7B (ft)~\cite{xia2024gsva} & 66.47 & 63.29 & 62.43 & 71.08 & 69.93 & 65.31 & 62.23 & 60.47 & 60.56 \\
    \rowcolor{mygray}
    AnchorSeg-LLaVA1.5-7B(ft) & \textbf{74.76} & \textbf{68.68} & \textbf{76.00} & \textbf{75.75} & \textbf{73.43} & \textbf{71.78} & \textbf{68.25} & \textbf{64.56} & \textbf{68.27} \\

    \bottomrule
    \end{tabular}}
    \label{tab:gres}
    \end{table*}

\section{Experiments}
\label{sec:experiments}

\subsection{Experimental Setting}
\label{exp:setting}

\paragraph{Implementation Details.}
We employ LLaVA1.5 (7B/13B)~\cite{liu2023llava} as the multimodal encoder $\mathcal{G}_{\mathcal{T}}$, while ViT-H SAM~\cite{kirillov2023segment} serves as the vision backbone $\mathcal{G}_{\mathcal{V}}$ specifically for mask generation. Visual features are extracted using CLIP-ViT-L/14@336, which operates on inputs of  size $336 \times 336$. To prevent object truncation caused by default center-cropping and to align the similarity map with SAM's input, we implement a padding-based resizing scheme (scaling the longest side to 336) and padding the image for the CLIP input. 
Training is performed on a single 
NVIDIA H800 GPU (80GB). We use a mixed dataset of ReasonSeg and referring segmentation samples, comprising approximately $\sim$10k images. 
We utilize LoRA~\cite{iclr2022lora} ($r=8$ for 7B; $r=64$ for 13B) and optimize via AdamW~\citep{loshchilov2017decoupled} with a learning rate of $3\text{e-}4$ and 100 warmup steps.

\paragraph{Evaluation Metrics.}
Consistent with established evaluation protocols in prior works~\citep{kazemzadeh2014referitgame, mao2016generation, lai2024lisa}, we adopt task-specific metrics. 
For reasoning segmentation, we report gIoU and cIoU. 
Specifically, gIoU is the mean Intersection-over-Union (IoU) computed per image, while cIoU is defined as the ratio of the cumulative intersection to the cumulative union over the dataset. 
For referring expression comprehension (REC), we use Precision@0.5, where a prediction is considered correct if its IoU with the ground truth is at least 0.5.

\subsection{Results on ReasonSeg Dataset}
\label{exp:reasonseg}

\paragraph{Comparison with the State-of-the-Art.}
In Table \ref{table:reason_seg}, AnchorSeg performs favorably against competitive baselines. On the ReasonSeg validation split, our 7B model achieves 68.3\% gIoU and 75.9\% cIoU, outperforming RSVP-GPT by +3.6\% and +12.8\%, respectively. This advantage carries over to the test set, where AnchorSeg surpasses RSVP-GPT by +4.3\% gIoU and +5.9\% cIoU on the overall split, with notable gains on long queries (+5.1\% gIoU and +8.8\% cIoU).   Scaling to 13B further improves performance. Compared to READ-13B, AnchorSeg achieves up to +5.5\% gIoU and +5.3\% cIoU on the test overall split, while delivering substantial improvements on long queries, validating its effectiveness in complex reasoning scenarios.

\subsection{Results on RefCOCO(+/g) Dataset}
\label{exp:referseg}
\paragraph{Comparison with the State-of-the-Art.} 

Table~\ref{table:refer_seg_c} presents the quantitative results for referring expression comprehension based on Precision@0.5. AnchorSeg consistently outperforms GSVA-Vicuna-7B across all evaluated benchmarks. On the RefCOCO dataset, AnchorSeg achieves 89.10\%, 92.10\%, and 84.49\% on the val, testA, and testB splits, respectively, surpassing the corresponding GSVA results. On RefCOCO+, AnchorSeg improves over GSVA by +8.08\%, +7.09\%, and +5.28\% on the val, testA, and testB splits. On RefCOCOg, AnchorSeg attains 84.09\% on val and 84.45\% on test, exceeding GSVA by +2.51\% and +2.62\%, respectively.

\subsection{Results on gRefCOCO Dataset}
\label{exp:gres} 
\paragraph{Comparison with the State-of-the-Art.} 

In Table~\ref{tab:gres}, we report results of AnchorSeg-LLaVA1.5-7B on the gRefCOCO dataset. Compared to the fine-tuned GSVA-Vicuna-7B, AnchorSeg achieves consistent improvements across all metrics. On the validation set, AnchorSeg achieves 74.76\% gIoU, 68.68\% cIoU, and 76.00\% N-acc., improving over GSVA by +8.29\%, +5.39\%, and +13.57\%, respectively. On Test A, AnchorSeg attains 75.75\% gIoU, 73.43\% cIoU, and 71.78\% N-acc., with gains of +4.67\%, +3.50\%, and +6.47\%. On Test B, AnchorSeg further improves over GSVA by +6.02\% gIoU, +4.09\% cIoU, and +7.71\% N-acc.

\begin{figure*}[h]
\vspace{-0.3cm}
\begin{center}
\includegraphics[width=1\linewidth]{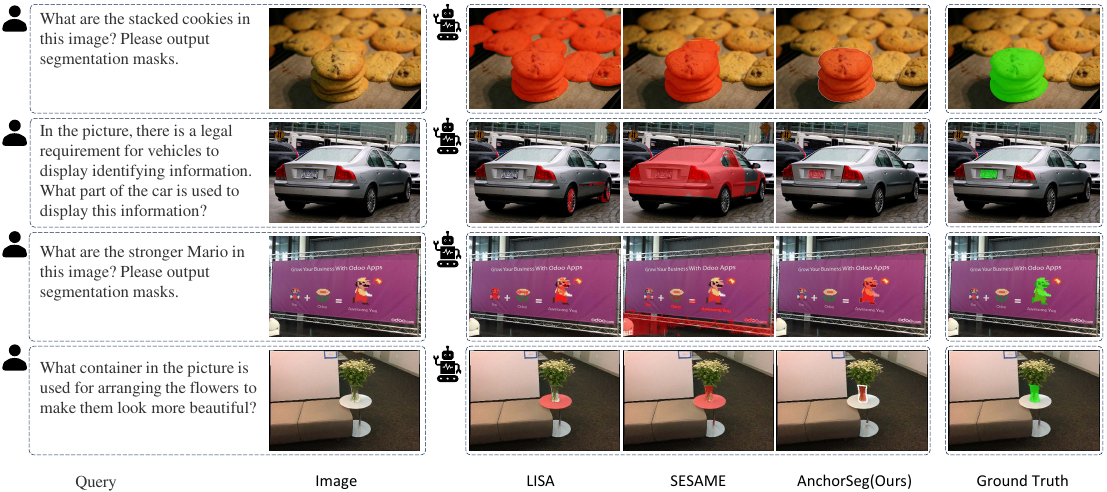}
\end{center}
\vspace{-0.4cm}
\caption{
Visual comparison of AnchorSeg (Ours) with prior works on the ReasonSeg \textit{val} set.
}
\label{fig:vis}
\vspace{-0.4cm}
\end{figure*}

\subsection{Qualitative Results}
\label{sec:vis}

Figure~\ref{fig:vis} shows that AnchorSeg achieves more precise segmentation results compared to prior methods such as LISA~\citep{lai2024lisa} and SESAME~\citep{wu2023see}. The model is particularly effective in fine-grained cases, as illustrated by the accurate segmentation of the target in the $4^{th}$ row.

\subsection{Ablation Study}
\label{sec:ablation_study}

In this section, we conduct ablation studies on the ReasonSeg \textit{val} set to investigate the contribution of each core component in our proposed AnchorSeg.

\begin{table}[h]
    \centering
    \caption{\textbf{Ablation on Conditional Query Sequence Scaling.} We report performance with varying sequence lengths $N$. Note that $N = K + 1$, comprising $K$ latent reasoning tokens and one anchor query \texttt{<SEG>}. $N=8$ yields the best performance.}
    \label{tab:ablation_scaling}
    \vspace{-0.3cm}
    \tabcolsep=0.4cm
    \resizebox{1\linewidth}{!}
    {
    \begin{tabular}{c|c|cc}
        \toprule
        Exp. ID & Sequence Len. ($N$) & gIoU & cIoU \\
        \midrule
        1 & 4 & 64.8 & 73.4 \\
        2 & 8 & \textbf{68.3} & \textbf{75.9} \\
        3 & 16 & 64.1 & 72.8 \\
        4 & 32 & 67.4 & 71.3 \\
        \bottomrule
    \end{tabular}
    }
\vspace{-0.3cm}
\end{table}

\paragraph{Impact of Conditional Query Sequence Scaling.}
We investigate the effect of scaling the conditional query sequence $\mathbf{Q}$. We vary the total length $N$ (where $N=K+1$, consisting of $K$ latent reasoning tokens and one anchor query \texttt{<SEG>}) to evaluate the representational capacity. As shown in Table~\ref{tab:ablation_scaling}, a short sequence ($N=4$, \ie, $K=3$) restricts the capacity to encode intermediate reasoning states, resulting in degraded performance (73.4\% cIoU). The performance peaks at $N=8$, achieving the best 68.3\% gIoU and 75.9\% cIoU, indicating an optimal balance between semantic reasoning and spatial grounding. However, further expansion ($N \ge 16$) leads to degradation, with cIoU dropping to 72.8\% at $N=16$, likely due to the increased optimization difficulty in the continuous embedding space.

\begin{table}[h]
    \centering
    \caption{\textbf{
    Ablation on the Contextual ($\boldsymbol{q}_{1:K}$) and Anchor ($\boldsymbol{q}_{anc}$) Queries, Spatial Prior $\mathbf{P}$, and TMCC.
    }}
\label{tab:ablation_query}
    \vspace{-0.3cm}
    \tabcolsep=0.25cm
    \resizebox{1\linewidth}{!}{
    \begin{tabular}{c|cccc|cc}
        \toprule
        Exp. ID & $\mathbf{P}$ & $\mathbf{TMCC}$  &$\boldsymbol{q}_{1:K}$ & $\boldsymbol{q}_{anc}$ & gIoU & cIoU \\
        \midrule
        1 & & & \checkmark & \checkmark & 51.8 & 61.4 \\
        2 & & \checkmark & \checkmark & \checkmark & 58.4 & 66.3 \\
        3 &\checkmark & & \checkmark & \checkmark & 68.3 & 71.6 \\
        4 & \checkmark &\checkmark &  & \checkmark & 67.5 & 74.0 \\
        5 & \checkmark &\checkmark & \checkmark & \checkmark & \textbf{68.3} & \textbf{75.9} \\
        \bottomrule
    \end{tabular}
    }
    \vspace{-0.5cm}
\end{table}

\begin{table}[h]
    \centering
    \caption{\textbf{Ablation on Token--Mask Cycle Consistency (TMCC).} We investigate the contribution of the cycle consistency terms: $\mathcal{L}_{T2M}$ and $\mathcal{L}_{M2T}$. The combination yields the best synergy.}
    \label{tab:ablation_tmcc}
    \vspace{-0.3cm}
    \tabcolsep=0.45cm
    \resizebox{1\linewidth}{!}
    {
    \begin{tabular}{c|cc|cc}
        \toprule
        Exp. ID & $\mathcal{L}_{T2M}$ & $\mathcal{L}_{M2T}$ & gIoU & cIoU \\
        \midrule
        1 & \checkmark & & 67.4 & 73.8 \\
        2 & & \checkmark & 64.4 & 68.2 \\
        3 & \checkmark & \checkmark & \textbf{68.3} & \textbf{75.9} \\
        \bottomrule
    \end{tabular}
    }
\vspace{-0.4cm}
\end{table}

\paragraph{Impact of  Language Grounded Conditioning.}
We progressively analyze the effects on factorization of contextual $\boldsymbol{q}_{1:K}$ and anchor $\boldsymbol{q}_{anc}$ queries, spatial prior $\mathbf{P}$, and TMCC. In Table~\ref{tab:ablation_query}, Exp. 1 denotes that neither the spatial prior nor TMCC is used during training. In this case, the contextual queries $\boldsymbol{q}_{1:K}$ and the anchor query $\boldsymbol{q}_{anc}$ have no functional differentiation and fully degrade into $\boldsymbol{q}_{1:K+1}$ \texttt{<SEG>} tokens, all serving as segmentation placeholder tokens (\ie, placeholders with no role-specific distinction). Exp. 2 denotes that both the spatial prior and TMCC are applied during training. However, the spatial prior is not provided to SAM (\ie, it is not added to the image feature). Exp. 3 incorporates the spatial prior $\mathbf{P}$ (derived from $\boldsymbol{q}_{anc}$) but omits TMCC, yielding a significant boost to 71.6\% cIoU. Exp. 4 applies both $\mathbf{P}$ and TMCC but discards the contextual queries $\boldsymbol{q}_{1:K}$, improving to 74.0\% cIoU. Finally, the full method (Exp. 5) re-introduces $\boldsymbol{q}_{1:K}$ to the decoder, achieving the highest 75.9\% cIoU, which demonstrates the effectiveness of the proposed design.

\paragraph{Impact of Token--Mask Cycle Consistency.}
We assess the effectiveness of the proposed Token--Mask Cycle Consistency (TMCC), which regularizes the latent spatial responses $\boldsymbol{S}$ via a bidirectional constraint. As shown in Table~\ref{tab:ablation_tmcc}, optimizing with only $\mathcal{L}_{T2M}$ yields 67.4\% gIoU and 73.8\% cIoU. In contrast, using $\mathcal{L}_{M2T}$ alone leads to a performance drop to 64.4\% gIoU, likely due to the coarse resolution of the latent space. By jointly optimizing both objectives ($\mathcal{L}_{TMCC} = \mathcal{L}_{T2M} + \mathcal{L}_{M2T}$), the model achieves the best performance, reaching 68.3\% gIoU and 75.9\% cIoU.

\begin{figure*}[htbp]
\vspace{-0.3cm}
\begin{center}
\includegraphics[width=1\linewidth]{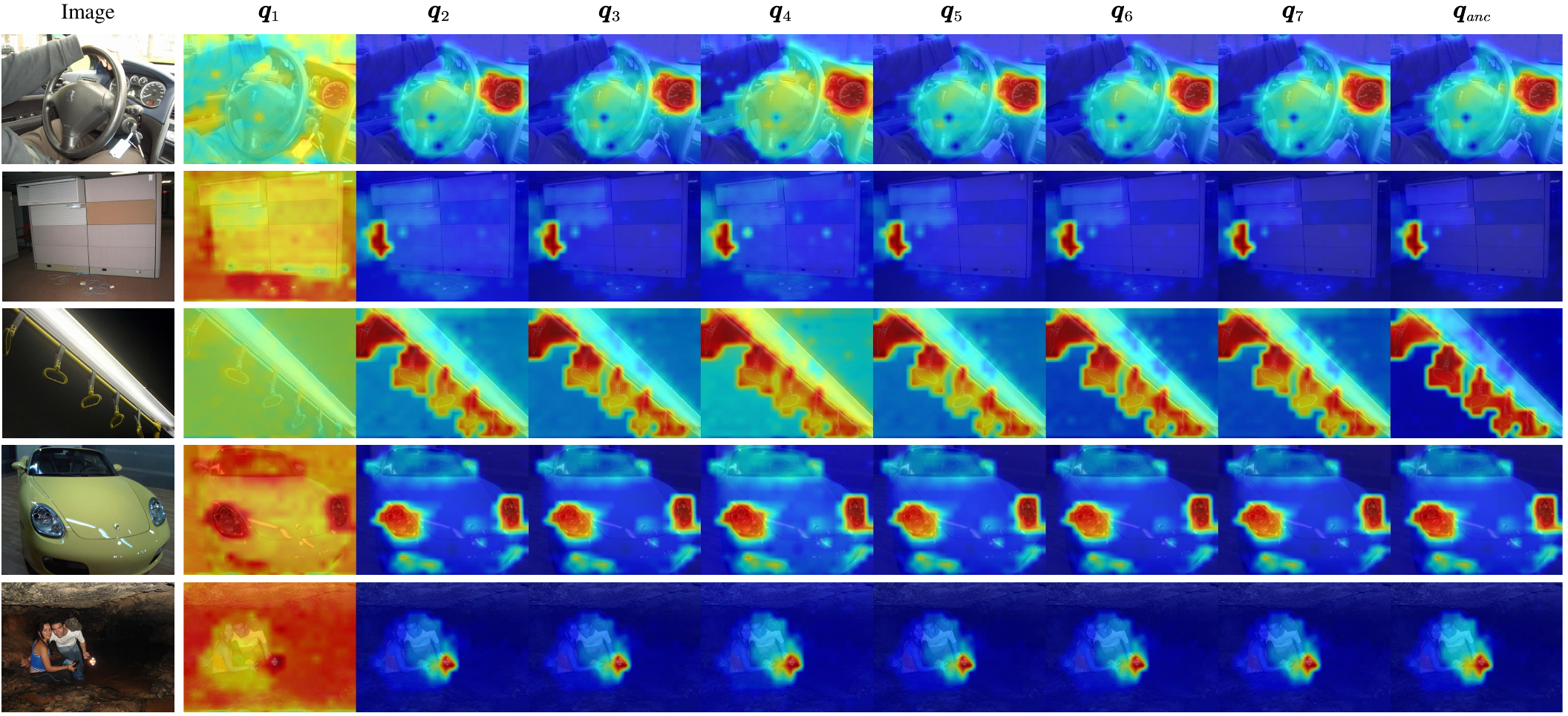}
\end{center}
\vspace{-0.4cm}
\caption{
Qualitative analysis of query banks. As the query tokens $\boldsymbol{q}_{1:7}$ approach the anchor token $\boldsymbol{q}_{anc}$, the activation gradually becomes less noisy and more spatially concentrated. The final anchor token produces a well-localized response, which serves as an effective spatial prior for guiding SAM segmentation.
}
\label{fig:query_banks}
\vspace{-0.2cm}
\end{figure*}

\begin{table*}[htbp]
\centering
\caption{Comparing the training cost of our AnchorSeg to state-of-the-art methods.}
\label{table:training_cost}
\vspace{-0.3cm}
\resizebox{1\linewidth}{!}
    {\tabcolsep=0.2cm
\begin{tabular}{lccccc}
\toprule
Model & Training Latency (s) & Memory Usage (GB) & Trainable (\%) & Trainable Params & Total Params \\
\midrule
SESAME~\cite{wu2023see} & 1.11 & 23.15 & 3.73\% & 288.25M & 7.73B \\
PixelLM~\cite{ren2024pixellm}    & 1.22 & 23.02 & 5.25\% & 375.72M & 7.16B \\
LISA~\cite{lai2024lisa}   & 1.26 & 23.68 & 3.74\% & 288.25M & 7.71B \\
GSVA~\cite{xia2024gsva}       & 2.13 & 25.73 & 3.73\% & 288.26M & 7.73B \\
AnchorSeg    & 2.94 & 29.89 & 3.94\% & 306.29M & 7.77B \\

\bottomrule
\end{tabular}
}
\vspace{-0.3cm}
\end{table*}

\begin{table*}[htbp]
\centering
\caption{Comparing the runtime speed of our AnchorSeg to state-of-the-art methods.}
\vspace{-0.25cm}
\label{table:runtime_speed}
\resizebox{1\linewidth}{!}
    {
    \tabcolsep=0.1cm
\begin{tabular}{lccccc}
\toprule
Model & GSVA~\cite{xia2024gsva} & SESAME~\cite{wu2023see} & LISA~\cite{lai2024lisa} & PixelLM~\cite{ren2024pixellm} & AnchorSeg \\
\midrule
Speed (FPS) & 3.98 & 4.64 & 4.68 & 9.24 & 4.00 \\
\bottomrule
\end{tabular}
}
\vspace{-0.5cm}
\end{table*} 

\subsection{Runtime Analysis} 
We evaluate the computational efficiency of AnchorSeg using a single NVIDIA A100-SXM4-40GB GPU, with results reported in Tables~\ref{table:training_cost} and~\ref{table:runtime_speed}. During training on the ReasonSeg dataset (batch size of 2), our method incurs an average iteration latency of 2.94 seconds and a peak memory usage of 29.89 GB. Compared to previous methods such as LISA~\citep{lai2024lisa} (1.26 s, 23.68 GB) and GSVA~\citep{xia2024gsva} (2.13 s, 25.73 GB), AnchorSeg exhibits a slightly higher training latency and memory usage, mainly due to the 
additional query 
construction and spatial conditioning modules. For inference performance (Table~\ref{table:runtime_speed}), AnchorSeg achieves a throughput of 4.00 FPS with batch size 1. This is comparable to GSVA~\citep{xia2024gsva} (3.98 FPS) and LISA~\citep{lai2024lisa} (4.68 FPS), while being slower than PixelLM~\citep{ren2024pixellm} (9.24 FPS). Overall, the results show that AnchorSeg maintains a reasonable trade-off between model complexity and runtime efficiency.  

\subsection{Query Analysis} 

We analyze the similarity maps of individual latent tokens to facilitate semantic interpretability in Fig.~\ref{fig:query_banks}. Considering a sequence length of $N=8$, the similarity map of $\boldsymbol{q}_1$ is highly noisy. For tokens $\boldsymbol{q}_{2:7}$, the noise gradually diminishes as they approach $\boldsymbol{q}_{anc}$, with the activation regions becoming increasingly concentrated. The activation corresponding to $\boldsymbol{q}_{anc}$ is well-localized. By leveraging the activation region of $\boldsymbol{q}_{anc}$ as a spatial prior to guide the downstream SAM, we observe a significant improvement in segmentation performance, as evidenced in Table~\ref{table:reason_seg}. This observation suggests a progressive refinement from coarse semantic responses to precise spatial grounding within the query sequence.
\section{Conclusion}
\label{sec: conclusion}

In this work, we reformulate reasoning segmentation as a structured conditional 
generation problem over image tokens, conditioned on language grounded query banks. 
We identify a key limitation in existing approaches: compressing both semantic 
reasoning and spatial localization into a single \texttt{<SEG>} token embedding 
restricts the model's ability to explicitly disentangle \emph{what to segment} 
from \emph{where to segment}. To address this, we introduce AnchorSeg, which
decouples spatial grounding from semantic reasoning through structured language grounded query banks. We hope our factorized formulation and 
explicit query bank design will inspire future research on language grounded visual 
perception and structured multimodal reasoning.

\section*{Acknowledgments} 
This work was supported by Dejing Dou’s Research Startup Fund
from Fudan University and the computations in this research were
performed using the CFFF platform of Fudan University.



\clearpage
\section*{Limitations}
\label{sec:limitations}

\noindent\textbf{Fixed Query Bank Size.} AnchorSeg uses a fixed number $K$ of latent 
reasoning tokens for all queries, regardless of reasoning complexity. However, simple 
queries (\eg, ``\emph{the red car}'') do not necessarily require multiple reasoning states, 
whereas more complex queries (\eg, ``\emph{the object that could prevent water damage 
in this scene}'') often benefit from richer intermediate reasoning. This mismatch suggests that adaptive mechanisms that dynamically allocate the number of latent tokens based on query complexity could improve both computational efficiency and overall performance. 

\bibliography{acl_latex}

\appendix

\clearpage
\setcounter{page}{1}

\appendix

\section{Ethics and Societal Impact}
\label{sec:ethics}

\textbf{Declaration of LLM Usage.} Large Language Models (LLMs) were used solely for linguistic refinement, including grammar checking and improving sentence fluency. In addition, LLMs were employed to assist with routine implementation details, such as data loading utilities and other boilerplate code. The conceptualization, methodological design, and experimental framework are entirely the original work of the authors. All AI-assisted outputs were carefully reviewed, verified, and edited by the authors to ensure correctness and consistency of the final manuscript.

\noindent\textbf{Broader Impact Statement.} This work is intended primarily for academic research purposes. We follow established ethical guidelines for data usage and ensure that all data processing complies with relevant protection regulations. All models used in this study are employed in accordance with their respective licensing agreements. To our knowledge, this work does not introduce direct negative societal impacts. We are committed to maintaining research integrity throughout the study.

\begin{table*}[t]
    \centering
    \caption{Overview of the training data mixture and its composition across different task categories.}
    \vspace{-0.1cm}\label{tab:dataset_mixture}
    \footnotesize
    \begin{tabular}{l l c c c c}
        \toprule
        \textbf{Dataset Types} & \textbf{Source Datasets} & \textbf{Total} & \textbf{Train Split} & \textbf{Val Split} & \textbf{Test Split} \\
        \midrule
        \multirow{5}{*}{sem\_seg} & ADE20K~\cite{zhou2017scene} & 22.21K & 20.21K & 2K & n.a. \\
                                & COCO-Stuff~\cite{caesar2018coco} & 118.29K & n.a. & n.a. & n.a. \\
                                & Pascal-Part~\cite{chen2014detect} & 4.37K & n.a. & n.a. & n.a. \\
                                & PACO-LVIS~\cite{ramanathan2023paco} & 45.79K & n.a. & n.a. & n.a. \\
                                & Mapillary~\cite{neuhold2017mapillary} & 18.00K & n.a. & n.a. & n.a. \\
        \cmidrule{1-6}
        
        \multirow{6}{*}{refer\_seg} & RefCLEF~\cite{kazemzadeh2014referitgame} & 17.98K & n.a. & n.a. & n.a. \\
                                  & RefCOCO~\cite{kazemzadeh2014referitgame} & 19.99K & 16.99K & 1.50K & 1.50K \\
                                  & RefCOCO+~\cite{kazemzadeh2014referitgame} & 19.99K & 16.99K & 1.50K & 1.50K \\
                                  & RefCOCOg~\cite{mao2016generation} & 25.80K & 21.90K & 1.3K & 2.6K \\
                                  & RefZOM~\cite{hu2023beyond} & 55.08K & 43.75K & n.a. & 11.33K \\
                                  & gRefCOCO~\cite{liu2023gres} & 19.99K & 16.99K & 1.50K & 1.50K \\
        \cmidrule{1-6}
        
        \multirow{3}{*}{neg\_refer\_seg} & R-RefCOCO~\cite{wu2024toward} & 15.10K & n.a. & n.a. & n.a. \\
                                       & R-RefCOCO+~\cite{wu2024toward} & 15.10K & n.a. & n.a. & n.a. \\
                                       & R-RefCOCOg~\cite{wu2024toward} & 20.04K & n.a. & n.a. & n.a. \\
        \cmidrule{1-6}
        
        \multirow{3}{*}{correct\_refer\_seg} & FP-RefCOCO~\cite{wu2023see} & 19.99K & 16.99K & 1.50K & 1.50K \\
                                           & FP-RefCOCO+~\cite{wu2023see} & 19.99K & 16.99K & 1.50K & 1.50K \\
                                           & FP-RefCOCOg~\cite{wu2023see} & 25.80K & 21.90K & 1.3K & 2.6K \\
        \cmidrule{1-6}
        
        vqa & LLaVA-Instruct-150K~\cite{liu2023llava} & 157.71K & n.a. & n.a. & n.a. \\
        \cmidrule{1-6}
        
        reason\_seg & ReasonSeg~\cite{lai2024lisa} & 1.22K & 239 & 200 & 779 \\
        \cmidrule{1-6}
        
        \multirow{4}{*}{\shortstack[l]{reason\_seg\_plus}} 
                                    & InstanceSeg~\cite{yang2023lisa++} & 58.50K & n.a. & n.a. & n.a. \\
                                    & CoT~\cite{yang2023lisa++} & 3.04K & n.a. & n.a. & n.a. \\
                                    & Conversations~\cite{yang2023lisa++} & 2.65K & n.a. & n.a. & n.a. \\
                                    & Caption~\cite{yang2023lisa++} & 1.34K & n.a. & n.a. & n.a. \\
        \cmidrule{1-6}
        
        \multirow{1}{*}{\shortstack[l]{multi\_reason\_seg}} 
                                    & MultiReasonSeg~\cite{ren2024pixellm} & 105.34K & 102.35K & 942 & 2.05K \\
        \bottomrule
    \end{tabular}
\end{table*}

\section{Datasets}
\label{sec:datasets}

Following prior data organization paradigms, we construct a comprehensive training mixture to support multi-granularity visual understanding and complex reasoning. In Table~\ref{tab:dataset_mixture}, we organize the training data into eight primary types: (1) \texttt{sem\allowbreak\_seg} for fundamental semantic understanding; (2) \texttt{refer\allowbreak\_seg} for standard referring expression comprehension; (3) \texttt{neg\allowbreak\_refer\allowbreak\_seg} and (4) \texttt{correct\allowbreak\_refer\allowbreak\_seg} for robust handling of false premises; (5) \texttt{vqa} for general conversational ability; and (6) \texttt{reason\allowbreak\_seg}, (7) \texttt{reason\allowbreak\_seg\allowbreak\_plus}, and (8) \texttt{multi\allowbreak\_reason\allowbreak\_seg} for advanced reasoning and multi-target grounding.

\textbf{Semantic Segmentation} (\texttt{sem\allowbreak\_seg}). This category includes several widely used datasets, such as ADE20K, COCO-Stuff, Pascal-Part, PACO-LVIS, and Mapillary. To align with the instruction-following objective, we randomly select target categories for each image and reformulate them as visual question-answering (VQA) pairs using predefined templates.

\textbf{Referring Segmentation} (\texttt{refer\allowbreak\_seg}). The RefCLEF dataset (\aka ReferItGame) is a widely used benchmark for referring expression comprehension, containing 130K expressions over nearly 100K regions across 20K images. In contrast to later COCO-based benchmarks, RefCLEF includes annotations for both discrete object instances and amorphous ``stuff'' categories.

The RefCOCO series (including RefCOCO, RefCOCO+, and RefCOCOg) comprises several large-scale benchmarks for referring expression comprehension and segmentation based on MS COCO. RefCOCO and RefCOCO+ contain relatively concise phrases, with RefCOCO+ further restricting absolute spatial terms to emphasize appearance attributes. In contrast, RefCOCOg features longer and more complex descriptions.

RefZOM is a challenging benchmark designed to address the one-to-one assumption in traditional referring segmentation, covering three settings: one-to-zero (no target), one-to-one (single target), and one-to-many (multiple targets). It contains 55,078 images and 74,942 annotated objects, split into 43,749 training images (58,356 objects) and 11,329 testing images (16,586 objects).

gRefCOCO is a large-scale benchmark for Generalized Referring Expression Segmentation (GRES), comprising 278,232 expressions, including 80,022 multi-target and 32,202 empty-target cases, over 19,994 images. The dataset is split into four subsets—training, validation, test-A, and test-B—following the UNC split of RefCOCO.

\textbf{Negative Referring Segmentation} (\texttt{neg\allowbreak\_refer\allowbreak\_seg}). The R-RefCOCO, R-RefCOCO+, and R-RefCOCOg datasets extend RefCOCO, RefCOCO+, and RefCOCOg by introducing negative text inputs that require empty-mask predictions. Following the original UNC and UMD splits, they augment the training data with a 1:1 positive-to-negative ratio and expand the validation sets with approximately 10 negative descriptions per reference object to evaluate robustness.

\textbf{Correct Referring Segmentation} (\texttt{correct\allowbreak\_refer\allowbreak\_seg}). FP-RefCOCO(+/g) extend the RefCOCO series for the false premise correction task. They use large language models to generate context-aware false premises by modifying objects, attributes, or relations. Following the original splits, each dataset maintains a 1:1 positive-to-negative ratio with approximately 20K--26K images to evaluate reasoning robustness.

\textbf{Visual Question Answering} (\texttt{vqa}). We use the LLaVA-Instruct-150K dataset, which contains approximately 158K vision-language instruction samples. It is included to maintain the model's general conversational ability and its capability to follow diverse natural language instructions.

\textbf{Reasoning Segmentation} (\texttt{reason\allowbreak\_seg}). ReasonSeg is a benchmark for reasoning segmentation, containing 1,218 image-instruction-mask samples. The images are annotated with implicit text instructions requiring complex reasoning or world knowledge, categorized into short phrases and long sentences. The dataset is split into 239 training, 200 validation, and 779 test samples.

\textbf{Reasoning Segmentation Plus} (\texttt{reason\allowbreak\_seg\allowbreak\_plus}). ReasonSeg-Plus is built on COCO 2017 and consists of four subsets: Instance Segmentation for instance-level segmentation, CoT (Chain-of-Thought) for global scene understanding and reasoning segmentation, Conversation for segmentation in multi-turn dialogue, and Caption for integrating segmentation with image captioning.

\textbf{Multi-target Reasoning Segmentation}  (\texttt{multi\allowbreak\_reason\allowbreak\_seg}). MUSE is a large-scale benchmark for multi-target reasoning segmentation, comprising 246K question-answer pairs and 910K instance-level mask annotations built on LVIS. The dataset features complex queries with arbitrary numbers of open-set targets, averaging 3.7 objects per response. It is split into 239K training, 2.8K validation, and 4.3K test samples to evaluate pixel-level reasoning and understanding.

\begin{table*}[!t]
\centering
\caption{Training recipe of AnchorSeg for different benchmarks.}
\vspace{-0.3cm}
\label{tab:recipe}
\renewcommand\arraystretch{1}
\small
\begin{tabular}{llcc}
\toprule
\multirow{2}{*}{\textbf{Benchmarks}} & \multirow{2}{*}{\textbf{Configuration}} & \multicolumn{2}{c}{\textbf{Model}} \\
\cmidrule(lr){3-4}
& & AnchorSeg-LLaVA1.5-7B & AnchorSeg-LLaVA1.5-13B \\
\midrule

\multirow{10}{2.3cm}{Reasoning Segmentation} 
& dataset & \multicolumn{2}{c}{\texttt{reason\_seg:refer\_seg:sem\_seg}=1:1:1} \\
& sem\_seg\_data & \multicolumn{2}{c}{\texttt{ade20k||cocostuff||pascal\_part||mapillary}} \\
& refer\_seg\_data & \multicolumn{2}{c}{\texttt{refclef||refcoco||refcoco+||refcocog||refzom||grefcoco}}\\
& epochs / steps\_per\_epoch & 120 / 10k & 30 / 10k\\
& grad\_accumulation\_steps& \multicolumn{2}{c}{ 10}
\\
& optimizer / lr & AdamW / $3\!\times\!10^{-4}$ & AdamW / $1\!\times\!10^{-4}$ \\
& betas / warmup\_num\_steps & \multicolumn{2}{c}{(0.9, 0.95) / 100 steps} \\
& lora\_r & 8 & 64 \\
& $\lambda_{bce}$ / $\lambda_{dice}$ & 2.0 / 4.0 & 2.0 / 4.0 \\
& zero\_stage & \multicolumn{2}{c}{2} \\
& num\_classes\_per\_sample & \multicolumn{2}{c}{3} \\
& batch\_size & \multicolumn{2}{c}{2 samples / GPU} \\
& weight\_decay & \multicolumn{2}{c}{0.00} \\
& gradient\_clipping & \multicolumn{2}{c}{1.0} \\
\midrule \addlinespace[0.5ex]

\multirow{10}{2.3cm}{Referring Expression Comprehension (REC)} 
& dataset & \multicolumn{2}{c}{\texttt{refer\_seg:sem\_seg:neg\_refer\_seg:correct\_refer\_seg=1:1:1:1}} \\
& refer\_seg\_data & \multicolumn{2}{c}{\texttt{refclef||refcoco||refcoco+||refcocog||refzom||grefcoco}} \\
& sem\_seg\_data & \multicolumn{2}{c}{\texttt{ade20k||cocostuff||pascal\_part||mapillary}} \\
& neg\_refer\_seg\_data & \multicolumn{2}{c}{\texttt{R-refcoco||R-refcoco+||R-refcocog}}
\\
& correct\_refer\_seg\_data & \multicolumn{2}{c}{\texttt{fprefcoco||fprefcoco+||fprefcocog}}
\\
& epochs / steps & \multicolumn{2}{c}{30 / 10k}
\\
& grad\_accumulation\_steps& \multicolumn{2}{c}{ 10}
\\
& optimizer / lr & AdamW / $3\!\times\!10^{-4}$ & AdamW / $1\!\times\!10^{-4}$\\
& lora\_r & 8 & 64 \\
& $\lambda_{bce}$ / $\lambda_{dice}$ & 2.0 / 4.0 & 2.0 / 4.0 \\
& zero\_stage & \multicolumn{2}{c}{2} \\
& num\_classes\_per\_sample & \multicolumn{2}{c}{3} \\
& batch\_size & \multicolumn{2}{c}{2 samples / GPU} \\
& weight\_decay & \multicolumn{2}{c}{0.00} \\
& gradient\_clipping & \multicolumn{2}{c}{1.0} \\
\midrule \addlinespace[0.5ex]

\multirow{10}{2.3cm}{Generalized Referring Expression Segmentation (GRES)} 
& dataset & \multicolumn{2}{c}{\texttt{refer\_seg}} \\
& refer\_seg\_data & \multicolumn{2}{c}{\texttt{refzom||grefcoco}} \\
& epochs / steps & \multicolumn{2}{c}{30 / 10k} \\
& grad\_accumulation\_steps & \multicolumn{2}{c}{10} \\
& optimizer / lr & AdamW / $3\!\times\!10^{-4}$ & AdamW / $1\!\times\!10^{-4}$\\
& lora\_r & 8 & 64 \\
& $\lambda_{bce}$ / $\lambda_{dice}$ & 2.0 / 4.0 & 2.0 / 4.0 \\
& zero\_stage & \multicolumn{2}{c}{2} \\
& num\_classes\_per\_sample & \multicolumn{2}{c}{3} \\
& batch\_size & \multicolumn{2}{c}{2 samples / GPU} \\
& weight\_decay & \multicolumn{2}{c}{0.00} \\
& gradient\_clipping & \multicolumn{2}{c}{1.0} \\
\bottomrule
\end{tabular}
\vspace{-0.5cm}
\end{table*}

\section{Implementation Details}
\label{sec:Details}

Given an input image $\mathbf{x}_{img} \in \mathbb{R}^{h \times w \times c}$ and a textual prompt $\mathbf{x}_{txt}$, LLaVA produces the image token embeddings $\boldsymbol{I}_{img} \in \mathbb{R}^{576 \times 4096}$ and the $\texttt{<SEG>}$ token embedding $\boldsymbol{q}_{anc} \in \mathbb{R}^{4096}$. The similarity map is computed as
\[
\boldsymbol{q}_{anc} \boldsymbol{I}_{img}^{T} \in \mathbb{R}^{1 \times 576},
\]
which is then reshaped to $1 \times 24 \times 24$, interpolated to $1 \times 256 \times 256$, and convolved to produce a spatial prior $\mathbf{P} \in \mathbb{R}^{256 \times 64 \times 64}$. The dense visual features from SAM are given by $\mathbf{f} = \mathcal{G}_{\mathcal{V}}^{enc}(\mathbf{x}_{img}) \in \mathbb{R}^{256 \times 64 \times 64}$. We inject the spatial prior via element-wise addition, yielding conditioned features $\tilde{\mathbf{f}} = \mathbf{f} \oplus \mathbf{P}$ for mask generation. The training recipes of AnchorSeg across different benchmarks are summarized in Table~\ref{tab:recipe}.

\subsection{Interpolation Details for Token--Mask Cycle Consistency (TMCC)}
\label{sup:tmcc_interp}

We specify the interpolation operators used in TMCC and refer to Algorithm~(\ref{alg:lgc}) for the complete token-to-image resizing pipeline.

\paragraph{Token-to-Mask Interpolation.}
Given token-level responses $\boldsymbol{S}\in\mathbb{R}^{N}$ with $N=G^2$ (\eg, $G=24$), the continuous similarity map used by the token-to-mask term $\mathcal{L}_{T2M}$ is constructed exactly following Algorithm~\ref{alg:lgc} (normalization, reshape to $G\times G$, bilinear upsampling to $L_{vl}=336$, aspect-ratio aligned crop, bilinear resize back to the supervision resolution).
In our code, all bilinear resizes are implemented via PyTorch {\small\texttt{F.interpolate(..., mode='bilinear')}}.
When resizing within the SAM-style preprocessing ({\small\texttt{apply\_image\_torch}}), we explicitly set {\small\texttt{align\_corners=False}} and {\small\texttt{antialias=True}}, and then pad zeros on the bottom/right to obtain a square canvas.
\begin{algorithm*}[htbp]
\caption{Language Grounded Conditioning Algorithm}
\label{alg:lgc}
\small
\begin{algorithmic}[1]
\REQUIRE
Input image $\mathbf{x}_{img}\in\mathbb{R}^{h\times w\times c}$ and text prompt $\mathbf{x}_{txt}$, 
vision-language model (\eg, LLaVA) that outputs contextual queries $\boldsymbol{q}_{1:K}\in\mathbb{R}^{K \times d}$ and anchor queries $\{\boldsymbol{q}^{(t)}_{anc}\in\mathbb{R}^{d}\}_{t=1}^{N_{seg}}$ (default $N_{seg}=1$).
For each $t$, the ordered query bank is $\mathbf{Q}^{(t)}=\big(\boldsymbol{q}_1,\dots,\boldsymbol{q}_K,\boldsymbol{q}^{(t)}_{anc}\big)$.
Image token embeddings $\boldsymbol{I}\in\mathbb{R}^{N\times d}$ (stacking $\boldsymbol{I}=\{\boldsymbol{i}_1,\dots,\boldsymbol{i}_N\}$ with $\boldsymbol{i}_j\in\mathbb{R}^{d}$).
Resize long-side target $L_{vl}=336$ for similarity alignment and $L_{sam}=256$ for SAM input. 
SAM visual features are given by $\mathbf{f} = \mathcal{G}_{\mathcal{V}}^{enc}(\mathbf{x}_{img}) \in \mathbb{R}^{C \times H \times W}$, with $(C,H,W)=(256,64,64)$. 
$f_{\theta}$ is a three-layer convolutional head with channel dimensions $1 \rightarrow4 \rightarrow 16 \rightarrow C$.
\ENSURE
Conditioned visual features $\tilde{\mathbf{f}}\in\mathbb{R}^{C\times H\times W}$ per anchor query $\boldsymbol{q}_{anc}$ for SAM mask generation.

\FOR{$t = 1,\dots,N_{seg}$}
    \STATE $\boldsymbol{s}^{(t)} \leftarrow \boldsymbol{I}\, \boldsymbol{q}^{(t)}_{anc}\in\mathbb{R}^{N}$ \qquad \qquad \qquad \qquad \qquad \qquad \quad  / * \texttt{Equivalently $s^{(t)}_j = \boldsymbol{i}_j^\top \boldsymbol{q}^{(t)}_{anc}$ for $j=1,\dots,N$} * /
    \STATE $\bar{\boldsymbol{s}}^{(t)} \leftarrow \dfrac{\boldsymbol{s}^{(t)}-\min(\boldsymbol{s}^{(t)})}{\max(\boldsymbol{s}^{(t)})-\min(\boldsymbol{s}^{(t)})+\epsilon}$
    \STATE $G \leftarrow \sqrt{N}$;\; $\tilde{\mathbf{P}}^{(t)} \leftarrow \mathrm{reshape}\!\left(\bar{\boldsymbol{s}}^{(t)};\, G\times G\right)$
    \STATE $\tilde{\mathbf{P}}^{(t)} \leftarrow \mathrm{interp}\!\left(\tilde{\mathbf{P}}^{(t)};\,L_{vl}\times L_{vl},\,\texttt{bilinear}\right)$
    \STATE $\alpha \leftarrow \dfrac{L_{vl}}{\max(h,w)}$;\; $h'\leftarrow \lfloor h\alpha + 0.5\rfloor$,\; $w'\leftarrow \lfloor w\alpha + 0.5\rfloor$
    \STATE $\tilde{\mathbf{P}}^{(t)} \leftarrow \mathrm{interp}\!\left(\tilde{\mathbf{P}}^{(t)}[\,:\,,\,0\!:\!h',\,0\!:\!w']^{(t)};\,h\times w,\,\texttt{bilinear}\right)$ \qquad \quad \quad /* \texttt{Crop and restore to ${h\times w}$.} */
    \STATE $\alpha \leftarrow \dfrac{L_{sam}}{\max(h,w)}$;\; $h'\leftarrow \lfloor h\alpha + 0.5\rfloor$,\; $w'\leftarrow \lfloor w\alpha + 0.5\rfloor$
    \STATE $\tilde{\mathbf{P}}^{(t)} \leftarrow 
\Big(
\mathrm{pad}(\,\cdot\,;\,(0,\,L_{sam}-w',\,0,\,L_{sam}-h')) 
\circ
\mathrm{interp}(\,\cdot\,;\,h'\times w',\,\texttt{bilinear})
\Big)\!\left(\tilde{\mathbf{P}}^{(t)}\right)$ \\ /* \texttt{Resize long side to $L_{sam}$, then right/bottom zero-pad to $L_{sam}\!\times\!L_{sam}$.} */
    \STATE $\mathbf{P}^{(t)} \leftarrow f_{\theta}\!\left(\tilde{\mathbf{P}}^{(t)}\right), f_{\theta}: \mathbb{R}^{{1 \times L_{sam} \times L_{sam}}} \rightarrow \mathbb{R}^{C\times H\times W}$
    \STATE $\tilde{\mathbf{f}}^{(t)} \leftarrow \mathbf{f}\oplus \mathbf{P}^{(t)}$
\ENDFOR
\RETURN $\{\tilde{\mathbf{f}}^{(t)}\}_{t=1}^{N_{seg}}$
\end{algorithmic}
\end{algorithm*}
\paragraph{Mask-to-Token Interpolation and Softening.}
For the reverse direction $\mathcal{L}_{M2T}$, the ground-truth mask $\mathbf{M}\in\{0,1\}^{H\times W}$ is first mapped to the $L_{vl}=336$ canvas using the same resize-and-pad operator as above (bilinear resize with {\small\texttt{align\_corners=False}}, {\small\texttt{antialias=True}}, followed by bottom/right zero padding), and then downsampled to the token grid using nearest-neighbor interpolation as
\begin{equation}
\scalebox{0.9}{
$\displaystyle
\begin{aligned}
\mathbf{M}^{grid} &= \mathrm{Interp}\!\left(\mathbf{M}^{336};\,G\times G,\,\texttt{nearest}\right), \\
\mathbf{M}^{grid} &\in \{0,1\}^{G\times G}.
\end{aligned}
$}
\end{equation}
The Gaussian soft target is then constructed on the token grid as 
\begin{equation}
\scalebox{0.9}{
$\displaystyle
\begin{aligned}
\mathbf{M}^{grid}_{\sigma} &= \mathbf{M}^{grid} * G_{\sigma}\in[0,1]^{G\times G}, \\
\mathbf{M}^{\downarrow}_{\sigma} &= \mathrm{vec}\!\left(\mathbf{M}^{grid}_{\sigma}\right)\in[0,1]^N.
\end{aligned}
$}
\end{equation}
We implement $G_{\sigma}$ using a normalized 2D Gaussian kernel with default $\sigma=7.0$ and nominal kernel size $31$,
with adaptive kernel sizing for small masks and {\small\texttt{reflect}} padding during convolution.

\subsection{Extending the Anchor Query to Multiple Spatial Anchors}
\label{app:multi_anchor}

While AnchorSeg uses a single segmentation anchor query $\boldsymbol{q}_{anc}$ to produce a language grounded spatial prior, the framework naturally extends to multiple spatial anchors to better support multi-target scenarios and more complex spatial relations.

\paragraph{Multi-target \& Multi-anchor Formulation.}
Let $N_{seg}$ denote the number of \emph{targets} to be segmented in an image (\ie, the number of masks to be predicted).
For each target $m\in\{1,\dots,N_{seg}\}$, the LMM can generate $T$ anchor tokens
$\{\texttt{<SEG>}_{m,t}\}_{t=1}^{T}$, yielding anchor queries
$\{\boldsymbol{q}^{(m,t)}_{anc}\}_{t=1}^{T}$ (optionally sharing the same contextual query bank $\boldsymbol{q}_{1:K}$).
For each anchor,  we compute token-to-image similarity as 
\[
s^{(m,t)}_i=\boldsymbol{i}_i^\top \boldsymbol{q}^{(m,t)}_{anc},
\]
producing spatial priors $\{\mathbf{P}^{(m,t)}\}_{t=1}^{T}$ for target $m$.
These priors can be injected into the SAM feature map either independently
(\eg, running the mask decoder $T$ times per target) or jointly via a lightweight fusion as 
\begin{equation}
\tilde{\mathbf{f}}^{(m)}=\mathbf{f}\oplus \Phi\!\left(\mathrm{concat}\left(\mathbf{P}^{(m,1)},\dots,\mathbf{P}^{(m,T)}\right)\right),
\end{equation}
where $\Phi(\cdot)$ is a small conv head. This yields one fused spatial prior per target and allows anchors to specialize to different regions within each instance.

\paragraph{When Multi-anchor Helps.}
We expect multiple anchors to be particularly beneficial for (i) explicit multi-object prompts (``segment \{A,B,C\}''); (ii) one-to-many referring and open-vocabulary multi-target segmentation; (iii) queries involving structured spatial relations (\eg, ``the two cups on the left of the plate''), where a single coarse spatial prior may be insufficient to disambiguate multiple valid regions.

\paragraph{Potential Challenges.}
Multi-anchor variants introduce additional computational overhead, which scales approximately linearly with $T$ when decoding is repeated. They may also lead to anchor competition and duplicate predictions. 
To mitigate these issues, practical strategies include anchor-wise non-maximum suppression (NMS) or merging to remove duplicate outputs, diversity regularization across anchors, and bipartite matching-based assignment during training when instance-level supervision is available. 
We leave the design of more efficient and principled multi-anchor generation and training strategies to future work.

\end{document}